\documentclass[conference]{IEEEtran}
\IEEEoverridecommandlockouts
\usepackage{cite}
\usepackage{multirow}
\usepackage{amsmath,amssymb,amsfonts}
\usepackage{algorithm}
\usepackage{algpseudocode}
\usepackage{graphicx}
\usepackage{textcomp}
\usepackage{xcolor}
\usepackage{mwe}
\usepackage{subfig}
\usepackage{booktabs}
\usepackage{float}
\floatstyle{plaintop}
\restylefloat{table}
\def\BibTeX{{\rm B\kern-.05em{\sc i\kern-.025em b}\kern-.08em
    T\kern-.1667em\lower.7ex\hbox{E}\kern-.125emX}}
    \usepackage{pgfplots}
\pgfplotsset{compat=1.15}
\usepackage{mathrsfs}
\usetikzlibrary{arrows}
\definecolor{wwwwww}{rgb}{0.4,0.4,0.4}
\definecolor{qqwuqq}{rgb}{0,0.39215686274509803,0}
\definecolor{qqqqff}{rgb}{0,0,1}
\definecolor{ccqqqq}{rgb}{0.8,0,0}
\makeatletter
\let\old@ps@headings\ps@headings
\let\old@ps@IEEEtitlepagestyle\ps@IEEEtitlepagestyle
\def\confheader#1{%
\def\ps@IEEEtitlepagestyle{
\old@ps@IEEEtitlepagestyle
\def\@oddhead{\strut\hfill#1\hfill\strut}
\def\@evenhead{\strut\hfill#1\hfill\strut}
}
\ps@headings
}
\makeatother
\confheader{
\small{Proceedings of the 10th RSI International Conference on Robotics and Mechatronics (ICRoM 2022), Nov. 15-18, 2022, Tehran, Iran} 
}
\usepackage[pscoord]{eso-pic}
\newcommand{\placetextbox}[3]{
\setbox0=\hbox{#3}
\AddToShipoutPictureFG*{ \put(\LenToUnit{#1\paperwidth},\LenToUnit{#2\paperheight}){\vtop{{\null}\makebox[0pt][c]{#3}}}
}
}
\placetextbox{.23}{0.055}{\textbf{\small{978-1-6654-2094-5/21/\$31.00~\copyright 2021 IEEE}}}
\begin{document}

\title{A Segment-Wise Gaussian Process-Based Ground Segmentation With Local Smoothness Estimation}

\author{\IEEEauthorblockN{ Pouria Mehrabi and Hamid D. Taghirad}
\IEEEauthorblockA{\textit{Advance Robotics \& Automated Systems
(ARAS)}\\
K. N. Toosi University of Technology, Tehran, Iran.\\
p.mehrabi@email and taghirad@.kntu.ac.ir}}
\maketitle
\begin{abstract}
Both in terrestrial and extraterrestrial environments, the precise and informative model of the ground and the surface ahead is crucial for navigation and obstacle avoidance. The ground surface is not always flat and in bumpy and rough scenes the functional relationship of the surface-related features may vary in different areas of the ground, and the measured point cloud of the ground does not bear smoothness. Thus, the ground-related features must be obtained based on local estimates or even point estimates. In this paper, a segment-wise $\mathcal{GP}$-based ground segmentation method with local smoothness estimation is proposed. The value of the length-scale is estimated locally for each data point which makes it much more precise for the rough scenes while being not computationally complex and more robust to under-segmentation, sparsity, and under-represent-ability issues. The segment-wise task is performed to estimate a partial continuous model of the ground for each radial range segment. Simulation results show the effectiveness of the proposed method to give a continuous and precise estimation of the ground surface in rough and bumpy scenes while being fast enough for real-time applications.
\end{abstract}


\section{Introduction}
In recent years there is an increased demand for different kinds of Unmanned Ground vehicles (UGVs), performing different tasks in a wide variety of applications. These increasing demands have given birth to various three-dimensional perception methods tailored for different conditions and task scenarios aiming at semantic aspects of the matter \cite{behley2019semantickitti, lim2021erasor}. The three-dimensional Light Detection and Ranging Sensor (LiDAR) has been extensively deployed for this purpose for its centimeter-level accuracy while being capable of measuring greater distances in comparison with cameras. Furthermore, LiDAR boasts multi-directional sensing ability \cite{Shin2017, Anders2016}. The captured three-dimensional point cloud from the LiDAR sensor is utilized for three main tasks: \textit{Tracking}, \textit{Detection} and \textit{Semantic Segmentation}. This paper is centered on the \textit{ground segmentation} task which is part of semantic segmentation. The ground segmentation task is often performed for two main goals: \textit{navigation} and \textit{semantic classification}. In the navigation mode, the algorithm tries to find the navigable areas to assist the UGV~\cite{Na2016, Byun2015} while in the semantic mode, the algorithm tries to precept neighboring objects to be able to track or avoid them. 

A precise ground segmentation not only will lead to a more precise object detection by methods like Euclidean clustering~\cite{Zermas2017} but also will reduce the computational cost of the further semantic segmentation tasks~\cite{Lim2022}. While being a necessary part of any perception module of UGVs, the ground segmentation methods still must deal with major challenges to be resolved. 

Every UGV maneuvering on the terrestrial surface inevitably encounters the ground. On flat surfaces, the ground may be easily estimated by filtering the point cloud using a height threshold or algorithms like  RANSAC~\cite{FischlerRANSAC} or \cite{Asvadi2016} which are simple plane fitting algorithms. While these methods are efficient to be performed for navigational goals in flat urban scenarios, they fail to be applicable where there exists a partially steep slope, bumpy sections, or a rough plateau in the navigable area. Furthermore, if applied for the perception goal they will cause under-segmentation in the final results~\cite{Narksri2018} where points related to the sloped or rough areas of the ground surface are taken into account as outliers or incorrectly merged with other objects.

Another challenge of the ground segmentation task is the sparsity of the data. As most of LiDAR's data is concentrated at near-origin locations, the data gets sparse at longer distances to the origin. Thus, the right ground plane is hard to be estimated at these locations. Available methods tend to neglect the sparse data and just estimate the ground for the near-origin locations which are insufficient for the perceptional goals. The same issue happens for the near-origin parts of the data while the ground plane's structure is far more complex to be estimated by partial or segment-wise methods, making the ground plane estimation methods bad representatives of the real ground plane. 

On the other hand, standard $\mathcal{GP}$ lacks the ability to adapt to the variable smoothness of the data. Thus, the input-dependent smoothness may not be taken into account by using the standard $\mathcal{GP}$ while they are essential in many applications such as ground segmentation. This arises from the fact that discontinuity and sudden changes in data density may happen in LiDAR's data, thus a regression method that is able to maintain its adaptivity to local features of the data has merit when compared with other methods. Different Methods have been proposed to estimate the structure of non-stationary covariance functions~\cite{Brooks2006ICRA}. The local smoothness can be modeled by a latent process and estimated using Markov Chain Monte Carlo approaches~\cite{Paciorek2003, Paciorek2003a}. Furthermore, some methods tend to obtain mean value estimates of the kernels~\cite{Plagemann2008},~\cite{YimingZhang2019}.

In Gaussian process-based ground segmentation the role of the length-scale value selection is shown to have a great impact on how the algorithm is able to predict the ground precisely~\cite{mehrabi2021gaussian},~\cite{Liu2019}. For the ground segmentation task~\cite{Chen2014} proposes a functional relationship of the form $\mathcal{L} = f(r)$ to relate the length scales to the input location. This is not sufficient because in non-flat grounds the functional relationship will not hold equitable for different regions of the ground.~\cite{Moosmann2009} assumes that length scales are a function of line features in different segments. This is not sufficient because no physical background is considered for the selection of functional relationships and this function might change and fail to describe the underlying data in different locations. \cite{mehrabi2021gaussian} tends to obtain physically motivated length-scale values by segmenting the point cloud into areas with probable common slopes using the notation of critical points. 

None of the methods above gives an exact mathematical framework to incorporate local characteristics of LiDAR`s non-smooth data. In this paper, a novel segment-wise method is proposed that estimates the continuous ground plane in a single segment using point estimates of smoothness. The input-dependent smoothness of the process is modeled using a latent stochastic process. Point estimates are obtained using the mean value prediction of the latent variables. The latent variables are assumed to be Gaussian distributed. Thus, point estimates of length-scale values at each particular ground candidate location are calculated using \textit{Maximum A Posteriori} (MAP) estimate of the latent variable. The gradient-based optimization is utilized to perform the maximization task. Modeling the input-smoothness of the data enables the method to locally adjust the regression model to the ground structure in a point-wise matter. This makes the method robust to under-segmentation as the regression task is adaptive to the local behavior of the elevation data. Furtherly, this makes the method robust to the sparsity issue by relying the height estimation on the point-wise characteristics in sections that there are few points to rely on. 

Furthermore, a pseudo-input selection algorithm is proposed to construct the support value set $\bar{\mathcal{D}} = (\bar{l}\in\bar{\mathcal{L}},\bar{r}\in\bar{\mathcal{R}})$ by collecting points from each \textit{line-segments} that are dividing the segment into different ranges based on local physical traits~\cite{mehrabi2021gaussian}. The pseudo-input selection process divides the area into different segments with different structures. Thus, in the near-origin location in which more complexity is present in the data, there would be more line segments. Thus, true representatives of the segments are chosen, making the method robust to the representability issue.  

Moreover, since for most UGV applications such as driver-less cars or most Autonomous Land Vehicle (ALVs), the behavior of the ground ahead is of greater importance than in other areas, the proposed method can be used in a single-segment setting to only estimate the ground ahead of the vehicle precisely and fast. The width of the area that segments include can be altered as a parameter of navigation strategy.  

\section{Radial Grid Map} 
A radial grid mapping is performed on LiDAR's three-dimensional point cloud. The point cloud data is segmented into $M$ different segments. Each segment is divided into different number of bins. The bin sizes are related to the radial range of the points in the corresponding segment.  
\begin{figure}[ht!]
\begin{center}
 \includegraphics[scale =0.2]{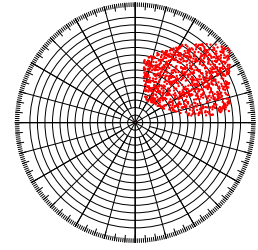}
 \label{fig:gridmap}
  \caption{A 2D Schematic of The Polar Grid Map}
  \end{center}
 \end{figure}
 The set of all the points in the $n_{th}$ bin of the $m_{th}$ segment is depicted by $P_{b_n^m}$ which covers the range $(r_n^{min}, r_n^{max})$. In order to reduce the computation effort the set $P_{b_n^m}^\prime$ is constructed that contains the corresponding two-dimensional points:
\begin{equation}\label{equation:four}
P_{b_n^m}^{\prime} = \{ p_i^{\prime} = (r_i,z_i)^T \; |\; p_i \in
P_{b_n^m}\},
\end{equation}
where, $r_i = \sqrt{x_i^2+y_i^2}$ is the radial range of corresponding points. The \textit{ground candidate set} $PG_m$, being the first intuitive guess for
obtaining an initial ground model is constructed by collecting the
the point with the lowest height at each bin as the ground candidate in
that bin \cite{Chen2014}:
\begin{equation}
PG_m = \{p_i^\prime \; | \; p_i^\prime \in P_{b_n^m}^{\prime}, z_i =
\min(\mathcal{Z}_n^m)\},
\end{equation}
where, $\mathcal{Z}_n^m$ is the set of height values in $P_{b_n^m}^\prime$.

\section{Nonlinear Gaussian Process Regression}
\section{Problem Formulation}
\subsection{The Predictive Distribution}
It is shown that the nonlinear Gaussian process regression problem is to
recover a functional dependency of the form $y_i = f (x_i)+\epsilon_i$
from $n$ observed data points of the training set $D$. In this
paper two different functional dependency is to recover: The
ground height $h_i = z(r_i) + \epsilon_{z_i}$ and the length-scale values $l_i = \mathcal{L}(r_i) + \epsilon_{l_i}$. The set $PG_m{(r_i, z_i)}$ contains all of the 2D ground candidate points in the segment $m$ that the ground model will be inferred based on.
The aim of the algorithm is to obtain the predictive ground
model $P(z^\ast | R^\ast , \mathcal{D},\theta)$:
\begin{eqnarray}\label{eq:intractablepredictive}
&P(z^\ast|R^\ast,\mathcal{D},\theta) =
\iint p\big(y^\ast|R^\ast,\mathcal{D},exp(\mathcal{L}^\ast),exp(\mathcal{L}),\theta_z\big) \nonumber \\ 
&\times p\big(\mathcal{L}^\ast,\mathcal{L}|R^\ast, R, \bar{\mathcal{L}},\bar{R},\theta_\mathcal{L}\big)dldl^\ast
\end{eqnarray}
In which $z^\ast$ is the regressed value at locations $\mathcal{R}^* \in \mathbb{R}^{1*d}$ given the data set $\mathcal{D}$ and hyper-parameters $\theta$. The parameter $d$ is the dimension of the prediction space. The parameter $\mathcal{R}$ represents the training data set. $\theta_z$ and $\theta_l$ represent the hyper-parameters. 

As the marginalization of the predictive distribution of Equation~(\ref{eq:intractablepredictive}) is intractable one method to approximate it is to apply Markov Chain Monte Carlo (MCMC)~\cite{Paciorek2003}. As \textit{Monte Carlo} methods are not efficient for the application, the most probable length-scale estimate is obtained: 
\begin{equation}
P(z^\ast|\mathcal{R}^\ast,\mathcal{D}, \theta) = P\big(z^\ast | \mathcal{R}^\ast,\mathcal{D}, exp(\mathcal{L}^\ast), exp(\mathcal{L}),\theta_z\big) 
\label{eq:predictive}
\end{equation}
The parameter $\mathcal{L}^\ast$ is the mean prediction of the length-scale value at input location $\mathcal{R^\ast}$ and $\mathcal{L}$ is the mean prediction of length-scale at the training data set's locations.
\subsection{Modeling The Input-Dependent Smoothness}
Length scales define the width of the area where observations may affect each other. The standard $\mathcal{GP}$ only considers a constant length scale over the whole input space. However, in applications like ground segmentation, observations may have different length scales. For example, on a large flat plateau height observations are strongly related to each other even in long distances while in a rough or bumpy section of the same elevation data, points may have no correlation to each other or have a very small impact on each other. To tackle this problem~\cite{Paciorek2003} has proposed the following non-stationary covariance function which incorporates the local characteristics at both locations to represent the correlation
of measured data points. 
\begin{eqnarray}\label{equation:nonscov}
&K(r_i,r_j) = \sigma_f^2.\big[\mathcal{L}_i^2\big]^{\frac{1}{4}}
\big[\mathcal{L}_j^2\big]^{\frac{1}{4}}
\big[\frac{\mathcal{L}_i^2 + \mathcal{L}_j^2 }{2} \big]^\frac{-1}{2}\nonumber \\
& \times  \exp\bigg(\frac{-(r_i-r_j)^2}{[  \mathcal{L}_i^2 +
\mathcal{L}_j^2]}\bigg),
\end{eqnarray}
Even by the selection of this covariance kernel for the $\mathcal{GP}$ regression, the problem of selecting a credible value for the length-scales $\mathcal{L}_i$ and $\mathcal{L}_j$ remains unresolved~\cite{Chen2014, Shen2021}. However, careful and credible selection of the length-scale values may lead to more credible results even for slopped terrains~\cite{mehrabi2021gaussian}. While a latent $\mathcal{GP}$ is put on the length scale values, their point-wise values are obtained using mean value estimation. Equation~\ref{equation:nonscov} is obtained by a mean value estimation
\subsection{The Case of Two Joint Gaussian Processes}
Two separate $\mathcal{GP}$s are assumed to model the ground segmentation task: $\mathcal{GP}_z$ and $\mathcal{GP}_l$. The $\mathcal{GP}_z$ is assumed to model the functional relation of the ground heights with the radial distance of the points with $(\mathcal{L}_*, \mathcal{L})$ treated as fixed parameters and $\mathcal{GP}_l$ is assumed to model the length-scales of the $\mathcal{GP}_z$s kernel function. The Predictive distribution of Equation~(\ref{eq:predictive}) enables the prediction of $z^\ast$ value at arbitrary locations $r^\ast$ at each point cloud frame. 
The Gaussian Process for Ground Height Values $\mathcal{GP}_z$ is defined as follows:
\begin{equation}
\mathcal{GP}_z: z(r) \sim \mathcal{GP}\big( \bar{z}(r),k(r,r^\prime)\big)
\end{equation}

With $z(r)$ being the height values at the location $r$, $\bar{z}(r)$ being the mean value of the height at location $r$ and $k(r,r^\prime)$ being the covariance kernel of Equation~\ref{equation:nonscov}. Predictive distribution of measurement process can be addressed only
after obtaining the line segments, support value set $\bar{\mathcal{D}}$ and consequently the local point estimate of length-scales. The predictive ground model of Equation~\ref{eq:predictive} enables estimation of height values $z^\ast$ at arbitrary locations $r^\ast$:
\begin{equation}
\mu_{\mathcal{GP}_z} = \bar{z} = K(r^*,r)^T\bigg[K(r,r)+
\sigma_n^2I\bigg]^{-1}z,
\end{equation}
\begin{equation}
\textrm{cov}_{z_*} = K(r_*,r_*) - K(r_*,r)\bigg[K(r,r)+
\sigma_n^2I\bigg]^{-1}K(r_*,r)^T.
\end{equation}
Furthermore, Matrix $\bar{K}(\bar{r},\bar{r})$ and vector
$\bar{K}(r,\bar{r})$ are corresponding co-variances for the latent process which is assumed to be a stationary, squared exponential covariance kernel:
\begin{equation}
\Rightarrow\bar{K}(\bar{r}_i,\bar{r}_j) =
\bar{\sigma}_f^2\exp\bigg(-\frac{1}{2}\frac{(\bar{r}_i -
\bar{r}_j)^2}{\bar{\sigma}_l^2}\bigg).
\end{equation}
Thus, by defining these two kernels, the predictive distribution of the height
$z^\ast$ at the arbitrary test input location $r^\ast$ is obtained using
the predictive ground model in which, $\mathcal{L}^\ast=\mu_{\mathcal{GP}_l}$ is the mean value prediction of the logarithm of the length-scale at the desired locations $\bar{R}^\ast$. 
\subsection{Pseudo-input Selection}
To obtain length-scale values for each data point, desired locations are adjusted to ground candidate data points at
each segment which are chosen based on the criteria that were defined in the previous work~\cite{mehrabi2021gaussian}. On the other hand, the $\bar{\mathcal{D}} = (\bar{l}\in\bar{\mathcal{L}},\bar{r}\in\bar{\mathcal{R}})$ which is the training set for the latent process must be obtained too. These locations of the test inputs for the latent kernel are chosen based on a \textit{pseudo input selection} algorithm which is performed to obtain these latent training data points.  

The \textbf{\textit{Physically-Motivated Line Extraction}} algorithm based on two-dimensional line fitting method is utilized to obtain \textit{\textbf{critical points}} in the segment. Then different \textbf{\textit{line-segments}} are constructed \cite{mehrabi2021gaussian}. Critical points are the points in which the behavior of the ground change due to different reasons such as a sudden change in the slope or a sudden change in the sparsity. As the critical points divide the segment into different successive bins with the same elevation behavior, the support values are chosen from these sets of bins in each line segment. To choose the support values from all the points in each line segment, the angular deviation of the points from the mean angular value of that line segment is evaluated which is depicted in Figure ~\ref{Fig:line-segment-pseudo}.

\begin{figure}
\begin{center}
\begin{tikzpicture}[scale = 0.5,line cap=round,line join=round,>=triangle 45,x=1cm,y=1cm]
\begin{axis}[
title = {Deviation from Mean Angular Value},
xlabel ={Height (z)},
ylabel = {Radial Distance (r)},
x=1cm,y=1cm,
axis lines=middle,
grid style=dashed,
ymajorgrids=true,
xmajorgrids=true,
xmin=0,
xmax=8,
ymin=0,
ymax=7,
xtick={-5,-4,...,12},
ytick={-3,-2,...,9},]
\clip(-5.0975,-3.03875) rectangle (12.8625,9.04125);
\draw [line width=0.8pt,color=ccqqqq] (1,2)-- (4.2825,4.92125);
\draw [line width=0.8pt,color=ccqqqq] (4.2825,4.92125)-- (6.5025,1.32125);
\draw [line width=0.8pt,color=ccqqqq] (6.5025,1.32125)-- (12.1825,6.86125);
\draw [->,line width=1.2pt,color=qqqqff] (1,2) -- (1,6);
\draw [->,line width=1.2pt,color=qqqqff] (1,2) -- (4,2);
\draw [->,line width=1.2pt,color=qqwuqq] (1,2) -- (4,5);
\draw [color=qqwuqq](1.0625,7.94125) node[anchor=north west] {The Mean Vector};
\draw [->,line width=1.2pt,color=wwwwww] (1,2) -- (2.4425,5.02125);
\draw [color=wwwwww](1.0825,8.56125) node[anchor=north west] {The Individual Points Vector};
\draw [color=ccqqqq](1.4625,7.78125) node[anchor=north west] {\parbox{2.8 cm}{ \textbackslash \textbackslash  Critical Points}};
\begin{scriptsize}
\draw [color=black] (1.4625,2.76125) circle (2.5pt);
\draw [color=black] (1.2425,2.24125) circle (2.5pt);
\draw [color=black] (1.7225,2.16125) circle (2.5pt);
\draw [color=black] (1.8825,2.58125) circle (2.5pt);
\draw [color=black] (2.4825,2.72125) circle (2.5pt);
\draw [color=black] (2.1425,2.98125) circle (2.5pt);
\draw [color=black] (2.3625,3.44125) circle (2.5pt);
\draw [color=black] (2.8425,3.58125) circle (2.5pt);
\draw [color=black] (3.2625,2.88125) circle (2.5pt);
\draw [color=black] (2.8225,3.16125) circle (2.5pt);
\draw [color=black] (3.0625,3.44125) circle (2.5pt);
\draw [color=black] (2.0825,3.84125) circle (2.5pt);
\draw [color=black] (1.7025,3.30125) circle (2.5pt);
\draw [color=black] (1.2625,3.52125) circle (2.5pt);
\draw [color=black] (1.5625,4.66125) circle (2.5pt);
\draw [color=black] (3.2225,3.76125) circle (2.5pt);
\draw [color=black] (3.8625,4.28125) circle (2.5pt);
\draw [color=black] (4.1825,3.20125) circle (2.5pt);
\draw [color=black] (4.6025,2.52125) circle (2.5pt);
\draw [color=black] (5.1225,2.42125) circle (2.5pt);
\draw [color=black] (4.5225,3.34125) circle (2.5pt);
\draw [color=black] (4.9425,3.38125) circle (2.5pt);
\draw [color=black] (5.9825,3.26125) circle (2.5pt);
\draw [color=black] (5.0225,3.64125) circle (2.5pt);
\draw [color=black] (5.5225,3.00125) circle (2.5pt);
\draw [color=black] (5.7625,2.00125) circle (2.5pt);
\draw [color=black] (5.4025,1.84125) circle (2.5pt);
\draw [color=black] (5.1225,2.18125) circle (2.5pt);
\draw [color=black] (4.8025,2.82125) circle (2.5pt);
\draw [color=black] (4.5025,3.76125) circle (2.5pt);
\draw [color=black] (5.6825,2.36125) circle (2.5pt);
\draw [color=black] (7.4425,2.96125) circle (2.5pt);
\draw [color=black] (6.4225,1.66125) circle (2.5pt);
\draw [color=black] (6.2825,2.14125) circle (2.5pt);
\draw [color=black] (6.5025,2.38125) circle (2.5pt);
\draw [color=black] (6.4625,3.06125) circle (2.5pt);
\draw [color=black] (6.3825,3.42125) circle (2.5pt);
\draw [color=black] (6.4825,3.90125) circle (2.5pt);
\draw [color=black] (6.9225,3.16125) circle (2.5pt);
\draw [color=black] (7.2625,2.32125) circle (2.5pt);
\draw [color=black] (8.4825,4.26125) circle (2.5pt);
\draw [color=black] (7.9225,4.42125) circle (2.5pt);
\draw [color=black] (7.7225,3.72125) circle (2.5pt);
\draw [color=black] (9.0825,2.44125) circle (2.5pt);
\draw [color=black] (8.5625,2.86125) circle (2.5pt);
\draw [color=black] (8.3625,3.18125) circle (2.5pt);
\draw [color=black] (8.8025,3.82125) circle (2.5pt);
\draw [color=black] (9.3225,3.54125) circle (2.5pt);
\draw [color=black] (9.2425,2.86125) circle (2.5pt);
\draw [color=black] (9.2025,2.08125) circle (2.5pt);
\draw [color=black] (11.1025,2.40125) circle (2.5pt);
\draw [color=black] (11.1225,2.74125) circle (2.5pt);
\draw [color=black] (10.3825,2.60125) circle (2.5pt);
\draw [color=black] (11.0425,2.34125) circle (2.5pt);
\draw [color=black] (10.8825,3.78125) circle (2.5pt);
\draw [color=black] (11.5225,4.76125) circle (2.5pt);
\draw [color=black] (11.1225,5.50125) circle (2.5pt);
\draw [color=black] (11.7425,6.10125) circle (2.5pt);
\draw [color=black] (10.6025,3.08125) circle (2.5pt);
\draw [color=black] (12.5625,4.70125) circle (2.5pt);
\draw [color=black] (12.1825,5.48125) circle (2.5pt);
\draw [color=black] (12,6) circle (2.5pt);
\draw [color=black] (12.0225,6.48125) circle (2.5pt);
\draw [color=ccqqqq] (4.2825,4.92125)-- ++(-2.5pt,-2.5pt) -- ++(5pt,5pt) ++(-5pt,0) -- ++(5pt,-5pt);
\draw [color=ccqqqq] (1,2)-- ++(-2.5pt,-2.5pt) -- ++(5pt,5pt) ++(-5pt,0) -- ++(5pt,-5pt);
\draw [color=ccqqqq] (6.5025,1.32125)-- ++(-2.5pt,-2.5pt) -- ++(5pt,5pt) ++(-5pt,0) -- ++(5pt,-5pt);
\draw [color=ccqqqq] (12.1825,6.86125)-- ++(-2.5pt,-2.5pt) -- ++(5pt,5pt) ++(-5pt,0) -- ++(5pt,-5pt);
\draw [color=ccqqqq] (1,6)-- ++(-2.5pt,-2.5pt) -- ++(5pt,5pt) ++(-5pt,0) -- ++(5pt,-5pt);
\draw [color=ccqqqq] (4,2)-- ++(-2.5pt,-2.5pt) -- ++(5pt,5pt) ++(-5pt,0) -- ++(5pt,-5pt);
\draw [color=qqqqff] (2.4425,5.02125)-- ++(-2.5pt,0 pt) -- ++(5pt,0 pt) ++(-2.5pt,-2.5pt) -- ++(0 pt,5pt);
\draw [color=ccqqqq] (1.1025,7.26125)-- ++(-2.5pt,-2.5pt) -- ++(5pt,5pt) ++(-5pt,0) -- ++(5pt,-5pt);
\end{scriptsize}
\end{axis}
\end{tikzpicture}
\label{Fig:line-segment-pseudo}
\caption{The black dots are the point clouds of the data. The red crosses and lines are representative of critical points and line segments. In each line segment, a mean vector is constructed (green arrow). Then in each segment, the difference in angle of each point's vector (the gray vector) is compared with this mean value to obtain the candidate support values.}
\label{Fig:line-segment-pseudo}
\end{center}
\end{figure}
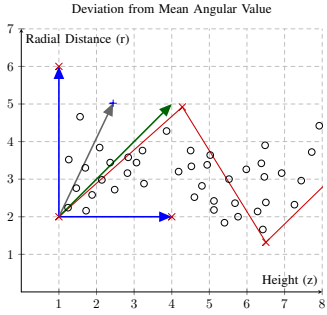

\subsection{Learning Hyper--Parameters}
 Until now, this paper's discussion was centered on the assumption that the hyperparameters have known values. However, acquiring \textit{a-priori} knowledge of the Hyperparameters is burdensome in many applications. Thus, the value for $\theta =
\{\sigma_f,\bar{\mathcal{L}}, \sigma_n, \bar{\sigma}_f,
\bar{\sigma}_l, \bar{\sigma}_n\}$ which are mutually independent variables, might be find by estimation of observations $z$. Thus, we tend to find the MAP estimate of the latent length scales, which means that we take the credible values the ones that maximize the likelihood of the length scales.  \emph{Log Marginal Likelihood:} Marginal likelihood or evidence
$P(\mathcal{L}|z,\mathcal{R},\theta)$ is the integral of likelihood times the prior. The logarithm of marginal likelihood is obtained under the Gaussian process
assuming that the prior is Gaussian $z|\mathcal{R}\sim
\mathcal{N}(0, K)$.

\begin{eqnarray}
&\log P(\mathcal{L}|z,R,\theta)=  \nonumber \\
&\log P(z|R,\exp(\mathcal{L}),\theta_z)+ \log
P(\mathcal{L}|R,\bar{\mathcal{L}},\bar{R},\theta_l)
\end{eqnarray}

\subsection{The Single--Segment Objective Function} The objective function is valid for each segment. In each segment, the structure of the kernels is different therefore different hyperparameters may be assumed for the latent kernel, thus the value for the objective function and its structure may vary. 
\begin{eqnarray}
J(\theta) & = & \log P(z^m|r^m,\theta) \\
 &  = &  -\frac{1}{2}\bigg[\bigg((z^m)^TA_m^{-1}z^m\bigg)+\log\bigg( |A_m|\bigg) \nonumber \\
 & + &\log\bigg( |B_m|\bigg)+\log(2\pi)\bigg((n_m+\bar{n}_m)\bigg)\bigg]\nonumber
\end{eqnarray}
where $A_m:=K_m(r,r)+\sigma_n^2I$ is the
corresponding covariance function of $\mathcal{GP}_y$ and $B_m:= \bar{K}_m(\bar{r},\bar{r}) + \bar{\sigma}_n^2I$ is the covariance matrix of $\mathcal{GP}_l$. The Scaled Conjugate Gradient optimization method is used for the optimization task~\cite{MOLLER1993525}.

\subsection{Gradient Evaluation}
The gradients of log marginal likelihood objective function
$L(\theta)$ are calculated analytically using,
\begin{eqnarray}
\frac{\partial L(\theta)}{\partial\ast} =
\bigg((z^m)^TA_m^{-1}
\frac{\partial A_m}{\partial \ast}A_m^{-1}z^m\bigg) \nonumber \\
-\frac{1}{2}\textbf{tr}
\bigg(A_m^{-1}\frac{\partial A_m}{\partial \ast}\bigg)
-\frac{1}{2}\textbf{tr}\bigg(B_m^{-1}
\frac{\partial B_m}{\partial \ast}\bigg)
\end{eqnarray}
It is then obvious that if we calculate the $\frac{\partial
A_m}{\partial \ast}$ and $ \frac{\partial B_m}{\partial \ast}$ for
all divided segments, the remaining calculations are found straight
forward. All the gradient calculations are detailed in~\cite{mehrabi2022probabilistic}.
\subsection{Evaluation Parameters}
To classify the test point, two attributes from~\cite{Douillard2011} are utilized to classify each input test point. First is the normalized distance between the real height value $z^\ast$ and the expected mean value of the height $\bar{z}$ at each test location which is depicted in the second plot of Figure~\ref{figure:overal}. The second attribute is the variance of the output $z^\ast$. A test point will be classified as a ground point if and only if satisfies the following inequalities at the same time:

\begin{equation}
    \frac{|z^\ast - \bar{z}|}{\sqrt{\sigma_n^2+\mathcal{V}_z}} \leq T_d \text{     and     }  \mathcal{V} \leq T_\mathcal{V}
\end{equation}
\section{Simulation and Results}
\begin{figure*}[ht]
    \centering
    \includegraphics[trim={1cm 0 1.5cm 1cm},clip,width=0.9\textwidth]{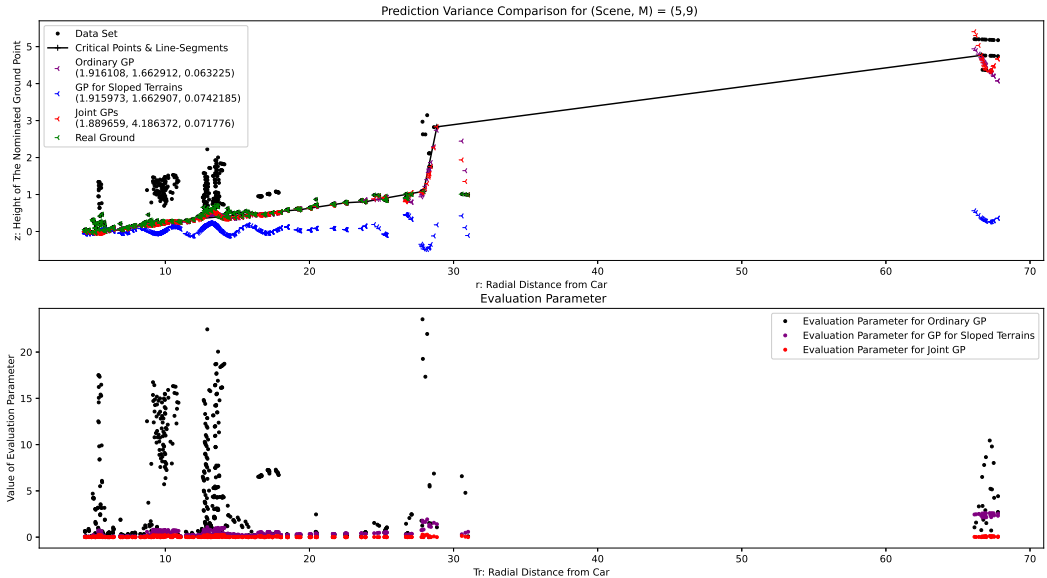}
    \caption{Estimation result of the proposed method reported with the estimation result of two other methods. The red dots are the mean value estimation $\bar{z}$ of the ground points in each test location. The Evaluation parameter of each method is reported too. It is depicted that the evaluation parameter of the proposed method is at the minimum in comparison with the other two methods, even in the sparse sections of the data $r\in(20m,33m)$ and even at the farthest locations.}
    \label{figure:overal}
\end{figure*}

The method is fully developed in C++. All modules and algorithms are implemented and developed from scratch using C++ in Qt Creator.
Furthermore, 3D point clouds are handled by the use of Point Cloud
Library (PCL). A Gaussian process regression ($\mathcal{GP}$) library
is also developed in C++ based on the math provided in
\cite{C.E.Rasmussen2006}. The Optimization method is developed in C++ using the \textit{alglib} library. 
\begin{table}[ht!]
\centering
\resizebox{\columnwidth}{!}{%
 \begin{tabular}{l c} 
 \hline
 Method & Average Run-Time(ms)\\ [0.5ex] 
\toprule
  Loopy Belief \cite{Zhang2015} & 1000\\
   GP-Based \cite{Chen2014}  & 200 \\
 Fast Ground Segmentation \cite{MinhChu2019b} & 94.7\\ 
 Slope-robust Cascade \cite{Narksri2018} & 76.511 \\
 GP-based for Sloped Terrains ~\cite{mehrabi2021gaussian} & 72.91\\
 GPF~\cite{Zermas2017} & 33.647 \\
 R-GPD (ERASOR)~\cite{lim2021erasor} & 28.328 \\
 PATCHWORK~\cite{Lim2022} & 22.742 \\
 Voxel-based \cite{Cho2014} & 19.31\\
 Fast segmentation \cite{Himmelsbach2010} & 16.96\\
 Enhanced Ground Segmentation \cite{MinhChu2019} & 6.9\\
 \textbf{\textit{Joint GP Method}} & \textbf{\textit{21.73}} \\
 \bottomrule 
 \end{tabular}
 }
     \caption{Average Run-Time of different ground segmentation methods.}
       \label{tableRT}
\end{table}
Data set in~\cite{Lai2009} is utilized to evaluate the results. To
present the efficiency of the method for under-segmentation and the sparsity issue, sparse segments and rough terrains are chosen. The results of the implementation of the algorithm are shown in Figure~\ref{figure:overal}. It
is depicted that the proposed method is giving a better estimation result for the sparse segments. Furthermore, it is depicted that the under-segmentation issue is handled by the algorithm. The average success rate of the proposed method is $96.57\%$ for sloped areas while the rate for flat areas is equal to $97.3\%$. The reported value from other methods is depicted in Table~\ref{tableSR}.
Furthermore, Table~\ref{tableRT} compares the processing time of different ground segmentation methods. Each frame of the data set presented by \cite{Lai2009} contains approximately 70,000 three-dimensional points. The Velodyne LiDAR sensor captures data at 10 $fps$. The reported average processing time of the proposed method is based on the implementation of an Intel Core i7-7700HQ CPU @ 2.80GHz. The proposed method is able to operate at 46 $fps$ which is about 4.6 times faster than the capture rate of the LiDAR sensors, making it suitable for real-time applications.
Furthermore, the run-time of the proposed method is compared to the run-time of the same methods in Table~\ref{tableRT}. It is depicted that the proposed method is not only real-time but among the fastest methods. This is because the method is performed segment-wise. Furthermore, The results prove the effectiveness of the proposed method to detect ground points even in rough scenes and while the data is sparse. 
\begin{table}[ht!]
\centering
\resizebox{\columnwidth}{!}{%
 \begin{tabular}{l c} 
 \hline
 Method & Average SR(flat/sloped)\\ [0.5ex] 
\toprule
 Fast Ground Segmentation \cite{MinhChu2019b} & 94.61/91.25\\ 
 Loopy Belief \cite{Zhang2015} & 97.19/NR\\
 Voxel-based \cite{Cho2014} & NR\\
 Enhanced Ground Segmentation \cite{MinhChu2019} & 94.71/91.70\\
 Fast segmentation \cite{Himmelsbach2010} & NR\\
 GP-Based \cite{Chen2014}  & 97.67 / NR\\
 GP-based for Sloped Terrains ~\cite{mehrabi2021gaussian} & 96.7/93.5\\
 \textbf{\textit{Joint GP Method}} & \textbf{\textit{97.3/96.57}} \\
 \bottomrule
  \end{tabular}
  }
     \caption{Average Success Rate (SR) of different ground segmentation methods.}
       \label{tableSR}
\end{table}

As depicted in Figure~\ref{figure:overal} the results show that the proposed algorithm is effective on the whole data without any data truncation or other pre-processings leading to data omission, the
advantage of which is the detection of the ground points which have greater heights or distance from the car. Finding the point estimates of the local smoothness has enabled the method to be accountable for the sparsity issue that happens in the LiDAR point clouds. For example, most of the ground segmentation methods in the literature only take near-vehicle data into account because far data is sparse. On the other hand, the \textit{represent-ability issue} is tackled by the introduction of the pseudo-input selection algorithm, which enables the method to base its estimation on the real physical condition of each segment.

\section{Conclusions}
A segment-wise ground segmentation method with point estimates of local smoothness is proposed that is capable of performing in rough and bumpy driving scenarios. The focus of the method is centered on tackling some fundamental issues of the ground segmentation task. The 3D LiDAR data is categorized based on a radial grid map to construct a segment-wise representation of 2D points. The Gaussian Process Regression $\mathcal{GP}$ is adapted to estimate the ground height values at each test location. It is mentioned that the Gaussian process regression ($\mathcal{GP}$) method fails to incorporate local smoothness into account even by using a non-stationary covariance. A mean value estimation strategy for length-scale estimation is utilized to make the $\mathcal{GP}$-based method adaptive to local attributes of the elevation data. The mean prediction strategy is adapted by introducing a second Gaussian process $\mathcal{GP}_l$ on the length-scale values. 

Moreover, the pseudo-input selection method is introduced to tackle the represent-ability problem while the local smoothness estimation enables the method to solve the sparsity issue of the 3D point clouds obtained from the LiDAR sensor. It is shown that the proposed method is able to be accountable in rough and bumpy scenes by solving the under-segmentation issue by being locally adaptive.

The method is tested on the labeled data set provided by~\cite{Lai2009}. The effectiveness of
the proposed method is verified by the results of the test which outperforms similar methods in rough scenes and in the overall determination of ground points even when no data truncation is performed, both for near-origin and far sparse data. 

\bibliographystyle{./bibliography/IEEEtran}
\bibliography{./bibliography/ALVGS}

\end{document}